\documentclass[sigconf,natbib=true]{acmart}
\usepackage{multirow,multicol}
\usepackage{amsmath,amsthm}
\usepackage{mathtools}
\usepackage{microtype}
\usepackage{subcaption} 
\usepackage{algorithm}
\usepackage{algorithmic}
\usepackage{fancyvrb}
\usepackage{graphicx}
\usepackage{makecell}
\usepackage{hhline}
\newcommand{\ours}{\textit{OpenDecoder}}

\AtBeginDocument{%
  }

\copyrightyear{2026}
\acmYear{2026}
\setcopyright{cc}
\setcctype{by}
\acmConference[WWW '26]{Proceedings of the ACM Web Conference 2026}{April 13--17, 2026}{Dubai, United Arab Emirates}
\acmBooktitle{Proceedings of the ACM Web Conference 2026 (WWW '26), April 13--17, 2026, Dubai, United Arab Emirates}
\acmPrice{}
\acmDOI{10.1145/3774904.3792524}
\acmISBN{979-8-4007-2307-0/2026/04}
\begin{document}

\title{OpenDecoder: Open Large Language Model Decoding to Incorporate Document Quality in RAG}

\author{Fengran Mo}
\orcid{0000-0002-0838-6994}
\affiliation{%
  \institution{Université de Montréal}
  \city{Montréal}
  \state{Québec}
  \country{Canada}
}
\email{fengran.mo@umontreal.ca}

\author{Zhan Su}
\orcid{0000-0001-5189-9165}
\affiliation{%
  \institution{Université de Montréal}
  \city{Montréal}
  \state{Québec}
  \country{Canada}
}
\email{zhan.su@umontreal.ca}

\author{Yuchen Hui}
\orcid{0000-0002-9659-3714}
\affiliation{%
  \institution{Université de Montréal}
  \city{Montréal}
  \state{Québec}
  \country{Canada}
}
\email{yuchen.hui@umontreal.ca}

\author{Jinghan Zhang}
\orcid{0009-0001-0999-270X}
\affiliation{%
  \institution{Clemson University}
  \city{Clemson}
  \state{South Carolina}
  \country{USA}
}
\email{jinghaz@clemson.edu}

\author{Jia Ao Sun}
\orcid{0000-0002-8340-155X}
\affiliation{%
  \institution{Université de Montréal}
  \city{Montréal}
  \state{Québec}
  \country{Canada}
}
\email{jia.ao.sun@umontreal.ca}

\author{Zheyuan Liu}
\orcid{0000-0001-7809-4586}
\affiliation{%
  \institution{University of Notre Dame}
  \city{Notre Dame}
  \state{Indiana}
  \country{USA}
}
\email{zliu29@nd.edu}

\author{Chao Zhang}
\orcid{0000-0003-3009-598X}
\affiliation{%
  \institution{Georgia Institute of Technology}
  \city{Atlanta}
  \state{Georgia}
  \country{USA}
}
\email{chaozhang@gatech.edu}

\author{Tetsuya Sakai}
\orcid{0000-0002-6720-963X}
\affiliation{%
  \institution{Waseda University}
  \city{Tokyo}
  \country{Japan}
}
\email{tetsuya@waseda.jp}

\author{Jian-Yun Nie}
\orcid{0000-0003-1556-3335}
\affiliation{%
  \institution{Université de Montréal}
  \city{Montréal}
  \state{Québec}
  \country{Canada}
}
\email{nie@iro.umontreal.ca}

\renewcommand{\shortauthors}{Fengran Mo et al.}

\begin{abstract}
  The development of large language models (LLMs) has achieved superior performance in a range of downstream tasks, including LLM-based retrieval-augmented generation (RAG).   
  The quality of generated content heavily relies on the usefulness of the retrieved information and the capacity of LLMs' internal information processing mechanism to incorporate it in answer generation. It is generally assumed that the retrieved information is relevant to the question. 
  However, the retrieved information may have a variable degree of relevance and usefulness, depending on the question and the document collection. It is important to take into account the relevance of the retrieved information in answer generation. 
  In this paper, we propose \ours{}, a new approach that leverages explicit evaluation of the retrieved information as quality indicator features for generation. We aim to build a RAG model that is more robust to varying levels of noisy context. Three types of explicit evaluation information are considered: relevance score, ranking score, and QPP (query performance prediction) score. 
  The experimental results on five benchmark datasets demonstrate the effectiveness and better robustness of \ours{} by outperforming various baseline methods. 
  Importantly, this paradigm is flexible to be integrated with the post-training of LLMs for any purposes and incorporated with any type of external indicators.
\end{abstract}



\maketitle

\section{Introduction}
The development of large language models (LLMs)~\cite{team2023gemini,zhao2023survey,achiam2023gpt} has achieved superior performance in a range of downstream tasks via their parametric knowledge acquisition from the training documents. 
However, LLMs still encounter foundational problems, such as understanding the limits of their knowledge and capability~\cite{heo2025llms,wang2025unveiling}, where a lack of sufficient knowledge might lead to hallucinations or generating outdated results~\cite{ji2023towards,li2024dawn}.
Retrieval-augmented generation (RAG)~\cite{lewis2020retrieval} is a common practice to address the incomplete knowledge issue by incorporating external information to obtain more accurate and reliable content generation.

Despite the fact that the RAG technique alleviates the knowledge boundary issue of LLMs, existing approaches to RAG face fundamental challenges: the quality of generated content heavily relies on the usefulness of the retrieved information and the capacity of LLMs' internal information processing mechanism~\cite{lin2025refrag,su2025parametric}. It is generally assumed that the retrieved information is relevant and useful for content generation, or LLMs have the capability to judge its relevance. However, the existing literature~\cite{du2022synthetic} showed the vulnerability of automated usefulness-checking systems when confronted with noisy information.
Thus, the defective and imperfect retrieved information would degrade the performance of LLMs. As a matter of fact, when an LLM is asked to answer a question based on an irrelevant document, the quality of the answer is negatively affected~\cite{tu2025robust}. Such a situation with irrelevant information may often occur when RAG is asked to deal with a large variety of questions. 
An ideal RAG system should be able to understand and tolerate the noisy input, i.e., process the diverse inputs that include useful evidence and irrelevant information, without being affected by the noise and resulting in significant degradation in performance~\cite{zhou2024trustworthiness,song2025measuring}.
For example, if the input context is partially noisy or extremely irrelevant, the system can attend only to the useful part or ignore the whole misinformation when generating an answer.

Existing studies attempt to address this issue from various perspectives, which can be categorized into (i) workflow-based methods and (ii) fine-tuning-based methods. 
The first category aims to design a workflow that navigates LLMs to identify useful pieces from retrieved information and append them to the final input context for generation.
The intermediate steps in the workflow vary and may include self-correction through LLM-as-a-judge~\cite{ye2024justice,gu2024survey}, isolating individual results for later aggregation~\cite{xiang2024certifiably,qian2025tackling}, and step-by-step filtering via reasoning~\cite{chang2024main}, among others.
This training-free approach is highly sensitive to the used prompt template and follows the strong assumption that the model could have enough capacity to distinguish the useful information by following the instruction to produce ideal output~\cite{heo2025llms}. 
However, one cannot expect that LLMs always generate correct judgments, and thus the manipulated final input might lose crucial information or include wrong information before conducting answer generation~\cite{yu2024rankrag}.
Besides, the judgment workflow with multiple steps with LLM calling would significantly increase  latency~\cite{csakar2025maximizing}.
On the other hand, the fine-tuning methods aim to teach the model to incorporate external useful knowledge in an effective way. For example, one can equip the LLMs with retrieval defect detection and utility extraction via instruction fine-tuning~\cite{tu2025robust,tang2025injecting} or enable the LLMs to interact with the retriever multiple turns until appending sufficient information for answer generation~\cite{asai2024self,jin2025search}.

Though effective, the existing approaches still inherit the original method of LLMs to perform the online computation of key-value pairs in the attention networks of the decoder~\cite{su2025parametric} for generation, which means that the autoregressive decoding of LLMs is mainly impacted by the attention score to produce generation probability.
We notice that the attention score is assigned by LLMs alone once the retrieved documents are appended into the prompt template. 
The original relevance judged by the retriever of the input documents is never used by the LLMs. 
Thus, the LLMs might treat the input documents as equally relevant or slightly different according to their input position~\cite{kim2025towards} based on the implicit internal judgments.
This gives rise to several critical questions: \textit{Should RAG ignore the relevance signals of the retrieved documents in its generation? Are such relevance signals useful for generation? How should the generation be impacted by document relevance?}

\begin{figure}[t]
\centering
\includegraphics[width=1\linewidth]{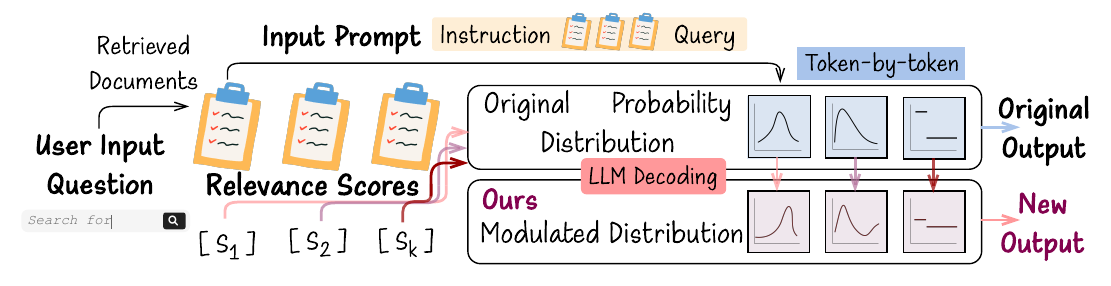}
\vspace{-4ex}
\caption{Comparison between the existing decoding LLMs that use their default probability distribution and our proposed approach that modifies the distribution by leveraging external explicit relevance signals.}
\label{fig: intro}
\vspace{-2ex}
\end{figure}

We believe that document relevance should be explicitly considered in answer generation in RAG, so that answer generation can be more tuned toward relevant information than irrelevant one.
To achieve this goal, in this paper, we propose \textbf{\ours{}}, a new approach that directly leverages document relevance to change the information processing procedure of LLMs decoding, namely, its attention mechanism.
As shown in Figure~\ref{fig: intro}, compared to the current decoding paradigm of LLMs, our proposed \ours{} does not only rely on the attention score produced via the internal network and instruction-following training, but also leverages explicit relevance signals as external indicator features.
The model is expected to become more robust to varying levels of noisy input context by reshaping the generation probability distribution via the useful information among the retrieved knowledge, and thus produce more accurate answers as output.

To implement \ours{}, the first step is to construct external indicators by extracting quality features from the retrieved documents.
We consider three types of signals: relevance score from the retriever, LLM-judged semantic score, and query performance prediction score. 
Then, we design a training framework to teach the LLMs to leverage these explicit indicator features (either separately or in combination) for answer decoding.
Specifically, we incorporate the external features into the internal attention networks computation to directly modulate the LLMs when producing generation probabilities for the decoding candidate tokens.
Additionally, to make the training and inference more robust to noisy information within the input, we conduct robustness training by reconstructing the input top-k documents via sampling additional documents with various relevant levels. 
During the online inference, the corresponding indicator features from external information are processed by the trained LLMs via the learned parameters in \ours{}.
Experiments on five benchmark datasets covering both general and multi-hop question answering (QA) demonstrate the effectiveness and enhanced robustness of the proposed approach, which consistently outperforms the vanilla RAG and other strong baselines across diverse noisy environments.
Importantly, our designed \ours{} is flexible to be integrated with the post-training of LLMs for any purposes and incorporate any other type of external indicator features towards effectiveness, robustness, or trustworthiness enhancement.

Our contributions are summarized as follows:

(1) We propose a new approach \ours{} to directly modify the LLM decoding in RAG by leveraging the relevance signals of 
the retrieved documents. 

(2) We design a training method, which includes constructing explicit relevance indicators from retrieved documents, teaching the model to leverage explicit indicators for answer decoding, and improving robustness via replacing the original top-k documents with various relevant levels ones.

(3) We conduct experiments on five widely used benchmarks, including general and multi-hop QA. Our \ours{} outperforms vanilla RAG and other strong baselines across diverse noisy environments, which demonstrates its superior effectiveness.

\section{Related Work}
\subsection{Retrieval-Augmented Generation}
Retrieval-Augmented Generation (RAG)~\cite{lewis2020retrieval,gao2023retrieval} aims to retrieve external resources to supplement 
LLMs to generate a response, showing significant advantages in knowledge-intensive tasks~\cite{guu2020retrieval,kang2023knowledge,dong2025decoupling}.
Earlier RAG methods follow the ``Retrieve-then-Read'' framework~\cite{lewis2020retrieval,izacard2023atlas} by adopting a retriever to search for relevant information from external resources based on the user's query. 
To further enhance RAG performance, subsequent studies focus on refining retrieval quality through techniques such as query reformulation~\cite{ma2023query,mo2023convgqr}, re-ranking~\cite{sun2023chatgpt,yu2024rankrag}, and noise filtering as intermediate steps~\cite{jin2024long,qian2025tackling}, thereby improving the relevance of documents before they are appended to LLMs' input.

However, retrieval errors remain common due to limitations in search effectiveness and corpus quality~\cite{petroni2020kilt}, which can ultimately degrade RAG performance.
To address this problem, robust RAG~\cite{liu2025robust,zhou2025trustrag} focuses on input optimization and knowledge integration. 
For instance, Weller et al.~\cite{weller2024defending} conduct query augmentation and introduce a novel confidence method based on answer redundancy.
RobustRAG~\cite{xiang2024certifiably} employs an isolate-then-aggregate strategy to ensure the robustness of LLM responses against retrieval corruption attacks. By generating self-synthesized rationales, InstructRAG~\cite{wei2024instructrag} explicitly denoises the retrieved content, thereby enhancing the robustness of RAG systems. AstuteRAG~\cite{wang2024astute} turns to refine and integrate knowledge derived from different sources to improve knowledge utilization and enhance the robustness of the generated answer.
RbFT~\cite{tu2025robust} proposes a robust fine-tuning strategy against retrieval defects with two defined tasks, defect detection and utility extraction, with associated instructions.
In addition, recent studies on developing deep search agents~\cite{jin2025search,li2025search,song2025r1,zheng2025deepresearcher} introduce a new paradigm for enhancing input quality by integrating in-context reasoning with dynamic search tool invocation when needed.
Although effective, these existing methods rely only on the internal mechanism of LLMs to process information, e.g., attention network~\cite{vaswani2017attention}.
Unlike them, our method \ours{} is developed to enable LLMs to distinguish useful information via both internal mechanisms and external explicit indicators.

\subsection{Decoding Optimization in LLMs}
Prompting~\cite{liu2023pre} the advanced LLMs is a simple and effective way to instruct them to generate answers, where the answer decoding highly relies on the designed prompt and internal attention mechanism.
Existing literature optimizes the decoding procedure of LLMs on various aspects.
For efficiency, Performers~\cite{choromanski2021rethinking} propose compressed attention, reducing attention complexity from quadratic to linear. StreamingLLM~\cite{xiao2024efficient} leverages attention sinks to decrease Key-Value cache memory for long-context generation. 
For effectiveness, a series of studies~\cite{chan2024rq,yue2025inference,tan2025rag} investigate how to leverage inference scaling and deep reasoning for RAG decoding.
A recent study REFRAG~\cite{lin2025refrag} rethinks RAG-based decoding and proposes an optimized architecture to compress only a small subset of retrieved documents that are directly related to the query for effective and efficient decoding.
For faithfulness, the existing studies aim to detect and manage misinformation within retrieved documents~\cite{zhou2024trustworthiness}, such as explicitly identifying and resolving knowledge conflicts~\cite{zhang2025faithfulrag,wang2025retrieval,deng2025cram}. These studies focus on selecting relevant and reliable information for LLM input, which still operate in the way that has been trained, by assuming the input information to be relevant. In contrast, in our approach, we modify the attention mechanism according to the relevance of retrieved information. Such an approach has not been proposed in the literature.

\section{OpenDecoder}
\begin{figure*}[t]
\centering
\includegraphics[width=0.95\linewidth]{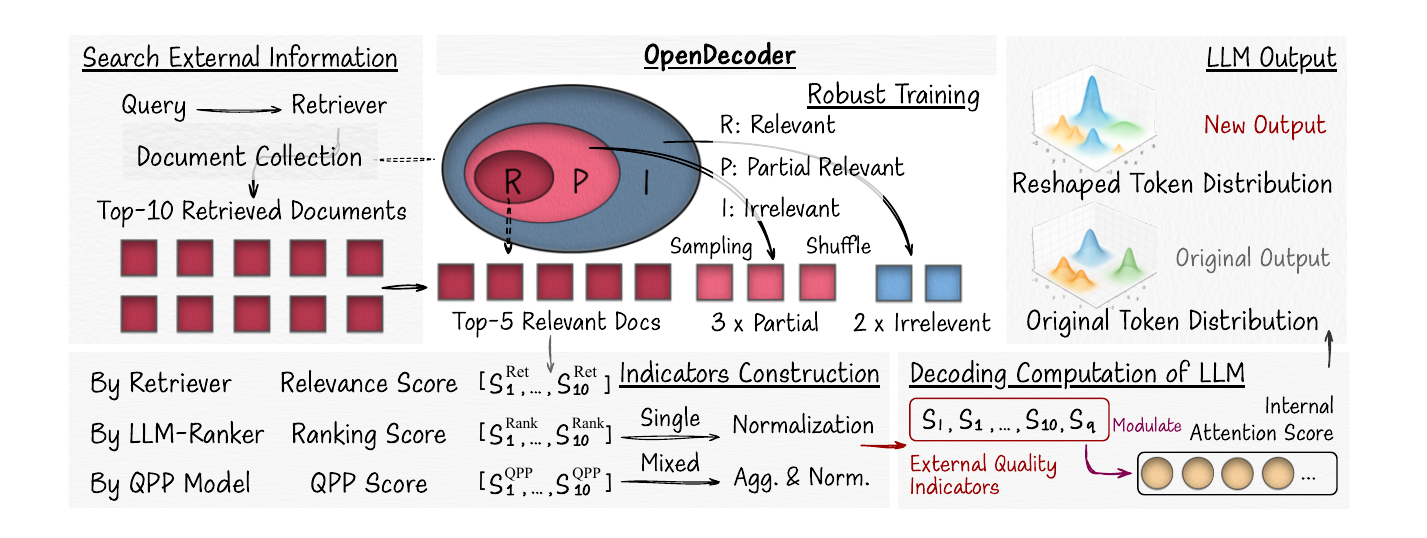}
\vspace{-2ex}
\caption{The framework of \ours{}, including Searching External Information with top-k retrieved documents, Indicators Construction based on the retrieved documents with various types of quality scores, teaching the model to leverage external explicit quality indicators for the Decoding Computation of LLM by modulating internal attention score computation and applying Robust Training, and finally obtaining the reshaped token probability distribution during content generation.}
\label{fig: overview}
\vspace{-2ex}
\end{figure*}

The principle of our methodology \ours{} is to modify the decoding procedure of LLMs with explicit relevance information as quality indicators, rather than solely based on prompt design. 
The goal is to enable the model to be robust to noisy retrieved information that can be irrelevant.
In the following sections, we first formulate the problem and provide an overview of our \ours{} as shown in Figure~\ref{fig: overview}.
Then, we present the detailed design for the components in \ours{}, including (1) constructing quality indicators via extracting features from external information; (2) learning to leverage explicit indicators for decoding; and (3) robustness training via replacing the input retrieved documents with various relevant levels.

\subsection{Task Formulation}
A vanilla RAG system typically consists of an ad-hoc retriever $\mathcal{R}$, a generator (i.e., the LLM) $\mathcal{G}$, and a corresponding corpus $\mathcal{C}$ with a large collection of 
documents. 
Given a user query $q$, the retriever $\mathcal{R}$ would  identify its top-k relevant documents $\mathcal{R}(q) = \{\text{doc}_i^q\}_{i=1}^k$. Then,  the LLM $\mathcal{G}$ would generates an answer $a$ based on the query and relevant documents 
as 
\raggedbottom
\begin{equation}
    a = \mathcal{G}(q, \{\text{doc}_i^q\}_{i=1}^k) = \mathcal{G}(q, \mathcal{R}(q, \mathcal{C}))
\end{equation}
\raggedbottom
The quality of the generated answer $a$ highly depends on the useful information returned by the retriever $\mathcal{R}$ and the understanding capacity of LLMs for the input context with the corresponding prompt. 
The inevitable noise in the retrieved context would significantly degrade the answer quality of LLMs on top of it.
These issues are unavoidable with the current prompting-based approach, where the content decoding only inherits the internal information processing mechanism of LLMs by following the prompt instruction~\cite{heo2025llms,wang2025unveiling}.
Our work focuses on guiding the decoding processing with explicit signals of external indicators of usefulness beyond the scores produced by the internal attention network.

\subsection{Constructing Indicators via Extracting Features from External Information}
\label{sec: Constructing Guidance Features}
Our goal is to incorporate external explicit indicators for LLMs to utilize internal knowledge stored in their parameters. Thus, the first step is to construct the indicators by extracting quality features from the retrieved information.
The most intuitive feature is the relevant score computed by the retriever model in terms of the given query and candidate documents.
In general, the retrieved top-k relevant documents $\{\text{doc}_i^q\}_{i=1}^k$ for the query $q$ are associated with their relevance scores $\mathcal{S}^\text{Ret}=\{s_i^{\text{Ret}}\}_{i=1}^k$, each computed by a similarity function as  $s_i^{\text{Ret}} = \frac{\text{q} \cdot \text{doc}_i^q}{\|\text{q}\| \, \|\text{doc}_i^q\|}$.
Since external indicators can be constructed in multiple ways, different features may be extracted and computed depending on the specific requirements, such as for faithfulness or trustworthiness.

In our implementation, we further leverage two additional indicators features, (i) the relevance judged by a LLM-based ranker as $\mathcal{S}^{\text{Rank}}=\{s_i^{\text{Rank}}\}_{i=1}^k$; and (ii) the query performance prediction (QPP) score $\mathcal{S}^{\text{QPP}}$ judged by a QPP model~\cite{meng2025query}. 
Specifically, we use the logit of the end-of-sequence token for the LLM-ranker judged score as $s_i^\text{Rank}= \text{Ranker}(\text{q}, \text{doc}_i^q)[-1]$ following~\citep{ma2024fine}, and the logit of token ``relevant'' in the prediction of the QPP model for each given document $\text{doc}_i^q$ in the candidate list as $s_i^\text{QPP}=\text{logit}\!\left(\text{``relevant''} \mid (q,\text{doc}_i^q) \right)$.
The relevance judged by the LLM-based ranker is expected to provide semantic similarity features from another perspective and help to investigate whether these explicit LLM-judged signals have additional impacts or have been integrated in model internal processing implicitly. Besides, the QPP scores provide the indicators about the difficulty of the query, which might imply the possible noisy level of the retrieved information for the generator.

Eventually, these scores calculated based on different aspects are used individually or as a combination $S^{\text{agg}}$ by an aggregation function to guide the LLMs to process the external information during generation, i.e., to decide to what extent it should focus on different parts of the input context in decoding.

\subsection{Learning to Leverage Explicit Indicators Features for Decoding}
\label{sec: Learning to Leverage Explicit Guidance Features for Decoding}
The fundamental problem in the current paradigm of RAG is that adding external retrieved information in the input prompt could only affect the online computation of key-value pairs in the attention networks of LLMs, which is not tailored to the input with noise. 
Since the retrieved context is usually not perfect, the inherent defects are only implicitly processed via the attention score computation, which is influenced by the mechanisms (e.g., predefined system prompt) in the pre-training procedure. 
Thus, a better way is to inform the decoding with additional explicit indicators directly, so that the LLMs know how much they should rely on external or internal knowledge to generate an answer.

To this end, we aim to teach the model to leverage the explicit indicator features from external information generated in Sec.~\ref{sec: Constructing Guidance Features}, and integrate them into the original attention networks computation.
Following the procedure of the standard RAG, the user query $q$ and its corresponding retrieved top-k documents $\mathcal{R}(q) = \{\text{doc}_i^q\}_{i=1}^k$ would fill the prompt template together with the instruction as $[\text{Instruction}, \text{doc}_1^q, \text{doc}_2^q, \cdots, \text{doc}_k^q, \text{query}]$ to instruct the LLM to produce an answer.
To teach the LLMs to leverage explicit indicator features, we first construct a score distribution by concatenating any types of score $\{s_i\}_{i=1}^k$ as features of the top-k retrieved documents and the pre-defined score $s_I$ and $s_q$ for the instruction $\mathcal{I}$ and query $q$ as $S = [s_I, s_1, s_2, \cdots, s_k, s_q]$. 
Then, we initialize it by normalizing the feature scores of the retrieved documents $\{s_i\}_{i=1}^k$ to $[0,1]$
and assign score $1$ to the tokens in query and instruction as Eq.~\ref{eq: norm}. 
The constructed score distribution $S_{\text{norm}} \in \mathbb{R}^{|S| \times |S|}$ is a token-level matrix, i.e., each token has an initial score value.
Finally, we incorporate the normalized scores $S_{\text{norm}}$ as explicit indicators into the computation of attention networks in \ours{} modified according to relevance as $\theta_{\text{open}}^{\text{attn}}$ via Eq.~\ref{eq: attention}.
The intuition is that, by modulating the original attention scores with normalized indicator scores, the importance of each token during the autoregressive decoding would be reshaped to guide the model for answer generation. 
In extreme cases where all input documents are irrelevant and assigned very low relevance scores, the query and instruction receive relatively higher scores, guiding the model to disregard the retrieved context and instead rely on its parametric knowledge to generate an answer.
\begin{algorithm}[t]
    \caption{Modulating LLM internal decoding in \ours{}}
    \label{alo: opendecoder} 
    \renewcommand{\algorithmicrequire}{\textbf{Input:}}
    \renewcommand{\algorithmicensure}{\textbf{Output:}}
    \begin{algorithmic}[1]
    \REQUIRE  Question $q$, 
    Relevance score $\{s_i\}_{i=1}^k$ of each input document, Normalization function $\text{Norm}(\cdot)$, Original $\mathcal{LLM}_{\theta_0}$.
    \ENSURE Updated $\mathcal{LLM}_{\theta_0+\theta_{\text{open}}^{\text{attn}}}$ and generated answer $a$.
    \STATE Normalize the relevance score among the input documents $\{s_i^{\text{norm}}\}_{i=1}^k = \text{Norm}(\{s_i\}_{i=1}^k)$.
    \STATE Construct token-level score matrix $S_{\text{norm}} \in \mathbb{R}^{|S| \times |S|}$ correspond to the input with question $q$ and instruction as Eq.~\ref{eq: norm}.
    \STATE Computation of modulated LLM's internal attention network $\mathcal{LLM}_{\theta_0}$ with external relevance score $S_{\text{norm}}$ via new parameter $\theta_{\text{open}}^{\text{attn}}$ as Eq.~\ref{eq: attention}.
    \STATE Generate a final answer $a$ for $q$ via the updated $\mathcal{LLM}_{\theta_0+\theta_{\text{open}}^{\text{attn}}}$.
    \end{algorithmic} 
\end{algorithm}
\raggedbottom
\begin{equation}
\label{eq: norm}
\begin{aligned}
    s_i^{\text{norm}} &= \frac{s_i}{\max (\{s_j\}_{j=1}^k)}, \quad s_q^\text{norm}, s_I^\text{norm} \leftarrow 1 \\
    S_{\text{norm}} &= [s_I^\text{norm}, \{s_j^{\text{norm}}\}_1^k, s_q^\text{norm}] \in \mathbb{R}^{|S| \times |S|}
\end{aligned}
\end{equation}
\raggedbottom
\begin{equation}
\label{eq: attention}
\theta_{\text{open}}^{\text{attn}} \sim \text{Attn}(Q, K, V, S_{\text{norm}}) = \text{softmax}\left(\frac{S_{\text{norm}} \cdot QK^{\top}}{\sqrt{d_k}} \right) V
\end{equation}
\raggedbottom
The type of scores and normalization approach can be determined according to various criteria such as relevance, reliability, authority, etc. 
In our implementation, we investigate three types of scores through an aggregation function before the normalization. We expect the relevance score $\mathcal{S}^{\text{Ret}}$ to be dominant and the other two scores $\mathcal{S}^{\text{Rank}}$ and $\mathcal{S}^{\text{QPP}}$ act as supplementary with a scale constant $0.5$, which is formulated in Eq.~\ref{eq: agg}.
\begin{equation}
\begin{aligned}
 \label{eq: agg}
    S_{\text{norm}}^{\text{agg}} &= \text{Normalize}\left(\text{Aggregate}(\mathcal{S}^{\text{Ret}}, \mathcal{S}^{\text{Rank}}, \mathcal{S}^{\text{QPP}})\right), \text{where} \\ s_{i-\text{agg}}^{\text{norm}} &= \frac{\left(s_{i-\text{Ret}}^{\text{norm}} + 0.5 * (s_{i-\text{Rank}}^{\text{norm}} + s_{i-\text{QPP}}^{\text{norm}})\right)}{\max (\{s_j^{\text{Ret}} + 0.5 * (s_j^{\text{Rank}} + s_j^{\text{QPP})}\}_{j=1}^k)},\
    s_{i-\text{agg}}^{\text{norm}} \in S_{\text{norm}}^{\text{agg}}
\end{aligned}
\end{equation}

Finally, we optimize to maximize the probability of producing the ground-truth $a$ with the given query and its corresponding retrieved top-$k$ documents set $\{\text{doc}\}_1^k$ as Eq.~\ref{eq: objective}, where $\theta_0$ and $\theta_{\text{open}}^{\text{attn}}$ denote the LLMs' original parameters and the learned parameters to leverage explicit quality indicator features during fine-tuning, respectively. During inference, the corresponding quality indicator features $\{s_i\}_{i=1}^k$ are required by learned parameters $\theta_{\text{open}}^{\text{attn}}$ for computation of probability in Eq.~\ref{eq: attention}. The core procedure of the information processing within the \ours{} is described in Algorithm~\ref{alo: opendecoder}.
\raggedbottom
\begin{equation}
\label{eq: objective}
    \max_{\theta} \sum_{(q,\{\text{doc}\}_1^k,a)} \sum_{t=1}^{|a|} \log \left(P_{\theta_0+\theta_{\text{open}}^{\text{attn}}}(a_t|a_{<t}, q, \{\text{doc}\}_1^k)\right)
\end{equation}
\raggedbottom

\subsection{Robustness Training}
\label{sec: Robustness Training}
It may often be the case that some retrieved documents are not relevant.
To make the training and inference more robust to 
noisy information, we conduct robustness training by replacing the second half of the top-k retrieved documents $\{\text{doc}_i\}_{i=1}^k$ with partial relevant ones $\{\text{doc}^{\text{part-rel}}\}$ and irrelevant ones $\{\text{doc}^{\text{irrel}}\}$ as Eq.~\ref{eq: robust}. They are sampled from the top-$k$ set excluding the top-5 documents and the whole collection excluding the top-$k$ documents, respectively.
The goal of constructing a noisy document list $\{\text{doc}\}_{\text{noisy}}$ is to provide a necessary environment for the model to learn to distinguish the useful and noisy information.
A further alternative is to shuffle the position of the noisy document list as $\{\text{doc}\}_{\text{noisy}}^{\text{shuffle}}$, aiming to emphasize the impact of external signals and reduce the common issue of position bias~\cite{gu2024survey,ye2024justice} of retrieved documents in RAG.
\begin{equation}
\label{eq: robust}
\begin{aligned}
    \{\text{doc}\}_{\text{noisy}} &= {\{\text{doc}_i\}}_1^5 \cup  \{\text{doc}^{\text{part-rel}}\} \cup \{\text{doc}^{\text{irrel}}\}, \quad \text{where} \\
    \{\text{doc}^{\text{part-rel}}\} &\sim \{\text{doc}_i\}_{i=6}^k, \{\text{doc}^{\text{irrel}}\} \sim (\mathcal{C}-\{\text{doc}_i\}_{i=1}^k)
\end{aligned}
\end{equation}
Then, the reconstructed noisy retrieved documents $\{\text{doc}\}_{\text{noisy}}$ or $\{\text{doc}\}_{\text{noisy}}^{\text{shuffle}}$ with various levels of noise and random relative position are used for robustness training by replacing the original input documents list $\{\text{doc}\}_1^k$ in Eq.~\ref{eq: objective}.

\begin{table*}[h]
\caption{Main results with three evaluation settings across various noisy environments among the retrieved documents for different RAG systems. To ensure fair and thorough comparison, all methods are based on Qwen-2.5-3B-Instruct backbone models, and the input retrieved documents for each method are fixed to the same. The best and second-best performance is set in \textbf{bold} and \underline{underline}.  $^\dagger$ and $^\ddagger$ denote significant improvements with t-test at $p<0.05$ over the strongest baseline RbFT and the Vanilla SFT without explicit external indicators for training, respectively. $\textasciicircum$/$^*$ represents in-domain/out-of-domain datasets.}
\vspace{-2ex}
\label{tab: main}
\centering
\resizebox{\textwidth}{!}{
\begin{tabular}{clcccccccccccc}
\toprule
\multirow{2}{*}{\bf Evaluation} & \multirow{2}{*}{\bf Method} & \multicolumn{2}{c}{\bf NQ$\textasciicircum$} & \multicolumn{2}{c}{\bf TrivialQA$^*$} & \multicolumn{2}{c}{\bf popQA$^*$} & \multicolumn{2}{c}{\bf HotpotQA$\textasciicircum$} & \multicolumn{2}{c}{\bf 2Wiki$^*$} & \multicolumn{2}{c}{\bf Average} \\
\cmidrule(lr){3-4} \cmidrule(lr){5-6} \cmidrule(lr){7-8} \cmidrule(lr){9-10} \cmidrule(lr){11-12} \cmidrule(lr){13-14}
& & F1 & EM & F1 & EM & F1 & EM & F1 & EM & F1 & EM & F1 & EM\\
\hline
\multirow{7}{*}{\bf Normal}
& No RAG & 12.11 & - & 30.04 & - & 11.07 & - & 16.86 & - & 22.38 & - & 18.49 & -\\
& Vanilla RAG & 25.46 & 34.12 & 31.09 & 48.40 & 7.35 & 21.87 & 12.06 & 20.73 & 11.23 & 20.93 & 17.44 & 29.21\\
& Vanilla SFT & 33.63 & 32.63 & 50.31 & 50.53 & 20.46 & 17.37 & 24.02 & 20.17 & 19.91 & 20.33 & 29.67 & 28.21\\
& RobustRAG & 26.58 & 30.80 & 45.25 & 47.40 & 10.91 & 15.90 & 13.58 & 15.73 & 5.84 & 9.07 & 20.43 & 23.78 \\
& InstructRAG & 30.39 & 31.00 & 45.81 & 50.73 & 16.19 & 21.70 & 20.26 & 22.93 & 17.37 & 20.87 & 26.00 & 29.45\\
& AstuteRAG & 37.84 & 34.10 & 52.28 & 51.80 & 23.92 & 19.90 & \underline{29.44} & 23.30 & 20.79 & 21.10 & 32.85 & 30.04\\
& RbFT & \textbf{40.17} & \textbf{36.60} & \underline{53.49} & \underline{52.30} & \underline{24.73} & \underline{21.42} & \textbf{29.71} & \textbf{24.50} & \underline{23.02} & \underline{21.90} & \underline{34.22} & \underline{31.34}\\
& \ours{} & \underline{39.26}$^\ddagger$ & \underline{35.90}$^\ddagger$ & \textbf{56.08}$^\dagger$$^\ddagger$ & \textbf{54.87}$^\dagger$$^\ddagger$ & \textbf{25.95}$^\dagger$$^\ddagger$ & \textbf{22.80}$^\dagger$$^\ddagger$ & \underline{29.44}$^\ddagger$ & \underline{24.00}$^\ddagger$ & \textbf{23.63}$^\ddagger$ & \textbf{22.53}$^\ddagger$ & \textbf{34.87}$^\ddagger$ & \textbf{32.02}$^\ddagger$\\
\hline
\multirow{6}{*}{\bf Noisy}
& Vanilla RAG & 15.22 & 32.70 & 26.82 & 49.93 & 7.83 & 20.66 & 11.05 & 19.00 & 11.97 & 20.38 & 14.58 & 28.53\\
& Vanilla SFT & 34.98 & \underline{32.83} & 48.54 & 48.07 & 21.06 & 18.16 & 23.55 & 20.80 & 22.07 & 20.40 & 30.04 & 28.05\\
& RobustRAG & 25.21 & 30.20 & 42.36 & 44.53 & 10.33 & 14.80 & 12.04 & 14.13 & 5.30 & 8.20 & 19.05 & 22.37 \\
& InstructRAG & 28.09 & 29.33 & 44.13 & 48.20 & 14.25 & 21.40 & 18.16 & 11.60 & 15.30 & 9.00 & 23.99 & 23.91\\
& AstuteRAG & 32.36 & 29.00 & 46.81 & 48.70 & 20.28 & 16.60 & 23.63 & 17.00 & 20.84 & 18.60 & 28.78 & 25.98\\
& RbFT & \underline{35.50} & 30.70 & \underline{52.62} & \underline{51.70} & \underline{23.71} & \underline{20.20} & \underline{25.28} & \underline{19.00} & \underline{23.60} & \underline{22.00} & \underline{32.14} & \underline{28.72}\\
& \ours{} & \textbf{37.71}$^\dagger$$^\ddagger$ & \textbf{33.82}$^\dagger$ & \textbf{55.09}$^\dagger$$^\ddagger$ & \textbf{53.33}$^\dagger$$^\ddagger$ & \textbf{25.07}$^\dagger$$^\ddagger$ & \textbf{22.02}$^\dagger$$^\ddagger$ & \textbf{28.76}$^\dagger$$^\ddagger$ & \textbf{22.77}$^\dagger$$^\ddagger$ & \textbf{24.17}$^\ddagger$ & \textbf{22.13}$^\ddagger$ & \textbf{34.16}$^\dagger$$^\ddagger$ & \textbf{30.81}$^\dagger$$^\ddagger$\\
\hline
\multirow{6}{*}{\bf Extreme}
& Vanilla RAG & 3.33 & 10.14 & 11.96 & 18.00 & 0.98 & 11.87 & 4.20 & 9.67 & 7.41 & 13.20 & 5.58 & 12.58\\
& Vanilla SFT & 19.78 & 16.73 & 34.76 & 33.40 & 19.27 & 18.37 & 18.26 & 15.07 & 21.76 & 19.93 & 22.77 & 20.70\\
& RobustRAG & 3.84 & 3.93 & 7.39 & 7.13 & 0.39 & 1.20 & 1.60 & 4.67 & 1.18 & 3.13 & 2.88 & 4.01\\
& InstructRAG & 5.52 & 7.40 & 21.51 & 24.80 & 1.62 & 0.70 & 9.14 & 5.80 & 11.25 & 6.80 & 9.81 & 9.10\\
& AstuteRAG & 16.06 & 9.50 & 35.03 & 27.10 & 15.74 & 12.80 & 14.38 & 10.60 & 17.36 & 15.10 & 19.71 & 15.02\\
& RbFT & \underline{21.49} & \underline{17.10} & \underline{38.18} & \underline{33.50} & \underline{21.59} & \underline{20.80} & \underline{22.11} & \underline{15.50} & \underline{24.28} & \underline{22.60} & \underline{25.53} & \underline{21.90}\\
& \ours{} & \textbf{22.50}$^\dagger$$^\ddagger$ & \textbf{18.06}$^\dagger$$^\ddagger$ & \textbf{40.41}$^\dagger$$^\ddagger$ & \textbf{38.27}$^\dagger$$^\ddagger$ & \textbf{24.96}$^\dagger$$^\ddagger$ & \textbf{22.02}$^\dagger$$^\ddagger$ & \textbf{23.59}$^\dagger$$^\ddagger$ & \textbf{17.20}$^\dagger$$^\ddagger$ & \textbf{26.99}$^\dagger$$^\ddagger$ & \textbf{24.00}$^\dagger$$^\ddagger$ & \textbf{27.69}$^\dagger$$^\ddagger$ & \textbf{23.91}$^\dagger$$^\ddagger$\\
\bottomrule
\end{tabular}}
\vspace{-1ex}
\end{table*}

\section{Experimental Setup}
\subsection{Datasets and Evaluation Metrics}
We evaluate \ours{} on five benchmark datasets, including two categories: (1) \textbf{General Question Answering}: NQ~\cite{kwiatkowski2019natural}, TriviaQA~\cite{joshi2017triviaqa}, and PopQA~\cite{mallen2023not}, and (2) \textbf{Multi-Hop Question Answering}: HotpotQA~\cite{yang2018hotpotqa} and 2WikiMultiHopQA~\cite{ho2020constructing}. These datasets encompass a diverse range of retrieval with noise in RAG, enabling a comprehensive evaluation in different settings.
Statistical details about the used datasets are provided in Appendix~\ref{app: Datasets}.

\subsection{Evaluation Settings in Noisy Environments}
We evaluate our \ours{} and all compared baselines among three settings with different noisy retrieval results. 
The first one is \textbf{Normal Evaluation}, where the input search results for RAG are the original top-10 documents from the retriever.
The second one is \textbf{Noisy Evaluation}, where the search results for RAG are constructed in the same way as the robust training in Sec.~\ref{sec: Robustness Training}, i.e., replacing the second half of the top-10 retrieved documents with partial relevant ones and irrelevant ones, which aims to evaluate whether the RAG system can distinguish the noise and solely rely on the useful input information.
The third one is \textbf{Extreme Noisy Evaluation}, where the search results for RAG are obtained by randomly sampling from the irrelevant document set, which simulates the extreme cases when the retrieval fails among difficult queries or domains. 

\subsection{Baseline}
To evaluate the effectiveness of \ours{} across various noisy settings, we compare it against the following baselines: 
(1) Vanilla retrieval-augmented generation (RAG)~\cite{lewis2020retrieval}; 
(2) Vanilla supervised fine-tuning (SFT)~\cite{chung2024scaling}; 
(3) RobustRAG~\cite{xiang2024certifiably}: An isolate-then-aggregate strategy to filter out the noise in retrieved context; 
(4) AstuteRAG~\cite{wang2024astute}: A retrieval-refined method to improve knowledge utilization and enhance robustness; 
(5) InstructRAG~\cite{wei2024instructrag}: Instructing LLMs to denoise retrieved content by generating self-synthesized explanatory rationales;
(6) Robustness fine-tuning (RbFT)~\cite{tu2025robust}: A more recent approach to conduct robustness training with two instruction fine-tuning tasks, defect detection and utility extraction.
More details about the baseline methods can be found in Appendix~\ref{app: Baseline}.

\subsection{Implementation Details}
We implement \ours{} based on Qwen-2.5 series backbone models~\cite{yang2025qwen3} with the official open-source code repository. 
The compared baselines are also implemented with the same Qwen-2.5-3B-Instruct model 
as our main experiments. 
For retrieval, we use the 2018 Wikipedia dump~\cite{karpukhin2020dense} as the knowledge source and E5~\cite{wang2022text} as the retriever, with the number of retrieved documents set to $10$, following~\cite{xiang2024certifiably,tu2025robust}.
For the robustness training, the number of relevant, partially relevant, and irrelevant documents is set to the same as the noisy evaluation, as 5, 3, and 2, respectively.
The partially relevant and irrelevant documents are randomly sampled five times from corresponding document sets and fixed for all compared methods for fair comparison.
For training, we merge the training sets of NQ and HotpotQA to form a unified training dataset for \ours{} and other fine-tuning-based baselines following~\cite{jin2025search}. The training epoch is set to 1 to ensure the model learn to use the explicit guidance and generalizes to out-of-domain evaluation datasets without overfitting.
Evaluation is conducted on the test sets of five datasets to assess both in-domain and out-of-domain performance. F1 score and Exact Match (EM) are used as the evaluation metrics, following~\cite{xiang2024certifiably,tu2025robust}. 
More implementation details can be found in our public code repository at \url{https://github.com/fengranMark/OpenDecoder}.

\section{Experimental Results}
\subsection{Main Results}
The overall performance of \ours{} is presented in Table~\ref{tab: main}. It is tested on five datasets, with three evaluation settings of different noisy environments in terms of the input retrieved documents. We can make the following observations:

(1) Our \ours{} consistently outperforms most compared baseline methods on three evaluation settings and significantly surpasses the Vanilla SFT approach without external indicators. Beyond the noisy and extremely noisy evaluation, the retrieved top-k documents in the normal evaluation might still contain noise in the input for answer generation (We will investigate the impact of noise Sec.~\ref{sec: Noise Tolerance of Input Top-K}). Thus, these results demonstrate the superior effectiveness of our \ours{} in tolerating noise, which can be attributed to our designed mechanism of modulating the decoding using external relevance signals as indicators and enabling the LLMs to grasp such capacity via specific training.

(2) Compared to other approaches targeting robustness improvement (RobustRAG and RbFT), our \ours{} exhibits more robust answer generation in noisy and extremely noisy settings. This is mainly because the compared methods still follow the current approach of internal information processing mechanism of the LLMs, which highly rely on the original capacity of the LLMs for distinguishing noise and the bias influenced by system prompts during pre-training. Modulating the LLM decoding with explicit indicators can not only provide useful signals but also alleviate this bias effect. 

(3) When the noise in the retrieved document increases, the performance drop is more severe in the relatively simple datasets (NQ and TrivialQA) compared with the other more complex ones (HotpotQA, 2wiki). This means the factoid questions with retrieved support evidence are more sensitive to the input with various noisy levels, thus the external indicators are more useful and necessary; while for the more difficult datasets, the retrieval defects are more common, and thus the urgent goal is to improve the success rate of retrieving relevant documents before aiming to enhance the robustness of the answer generation.

\subsection{Ablation Study}
The ablation studies are shown in Table~\ref{tab: ablation}. We can observe that by providing explicit indicators, the LLMs can better process the input information compared to the Vanilla SFT, which is the key idea of our method, and achieve the highest improvement. The feature aggregation is effective in some datasets, while robust training can contribute more to stable performance.
A possible explanation is that the most effective features may vary depending on the distribution of the dataset, necessitating a more adaptive feature selection mechanism to enhance generalizability. Meanwhile, introducing noisy training inputs remains essential to improve the model's robustness and noise tolerance during inference.
Nevertheless, combining all these mechanisms for implementing \ours{} can obtain better results across three different evaluation settings with various levels of noisy context on five datasets, which indicates the effectiveness of each component.

\begin{table}[t]
\caption{Ablation studies on the effectiveness of each mechanism in our \ours{} training framework.}
\vspace{-2ex}
\label{tab: ablation}
\centering
\scalebox{0.9}{
\begin{tabular}{lccccc}
\toprule
\multirow{1}{*}{\bf Method} & \multicolumn{1}{c}{\bf NQ} & \multicolumn{1}{c}{\bf TrivialQA} & \multicolumn{1}{c}{\bf popQA} & \multicolumn{1}{c}{\bf HotpotQA} & \multicolumn{1}{c}{\bf 2Wiki}
\\ 
\hline
\multicolumn{6}{l}{\bf Normal Evaluation} \\
Vanilla SFT & 33.63 & 50.31 & 20.46 & 24.02 & 19.91 \\
w/. Guidance & 37.62 & 55.31 & 24.33  & 26.06 & 20.15 \\
w/. Aggregate & 36.24 & 55.48 & 21.59 & 28.86 & 22.85 \\
w/. Robust Tr. & 38.98 & 55.84 & 25.14 & 29.43 & 22.72\\
\ours{} & 39.26 & 56.08 & 25.95 & 29.43 & 23.63\\
\hline
\multicolumn{6}{l}{\bf Noisy Evaluation} \\
Vanilla SFT & 34.98 & 51.37 & 21.06 & 23.55 & 22.07\\
w./ Guidance & 37.30 & 53.35 & 24.05 & 25.78 & 23.65\\
w/. Aggregate & 36.42 & 53.84 & 23.96 & 28.39 & 23.38\\
w/. Robust Tr. & 37.43 & 54.57 & 24.56 & 28.39 & 23.33\\
\ours{} & 37.71 & 55.09 & 25.07 & 28.76 & 24.17\\
\hline
\multicolumn{6}{l}{\bf Extreme Noisy Evaluation} \\
Vanilla SFT & 19.78 & 34.76 & 19.27 & 18.26 & 21.76 \\
w/. Guidance & 21.89 & 39.03 & 24.58 & 20.28 & 22.79 \\
w/. Aggregate & 21.07 & 39.57 & 24.26 & 23.25 & 26.77\\
w/. Robust Tr. & 22.22 & 40.33 & 25.61 & 23.36 & 26.52 \\
\ours{} & 22.50 & 40.41 & 24.96 & 23.59 & 26.99\\
\bottomrule
\end{tabular}}
\vspace{-2ex}
\end{table}

\begin{figure*}[t]
\centering
\vspace{-2ex}
\includegraphics[width=0.97\linewidth]{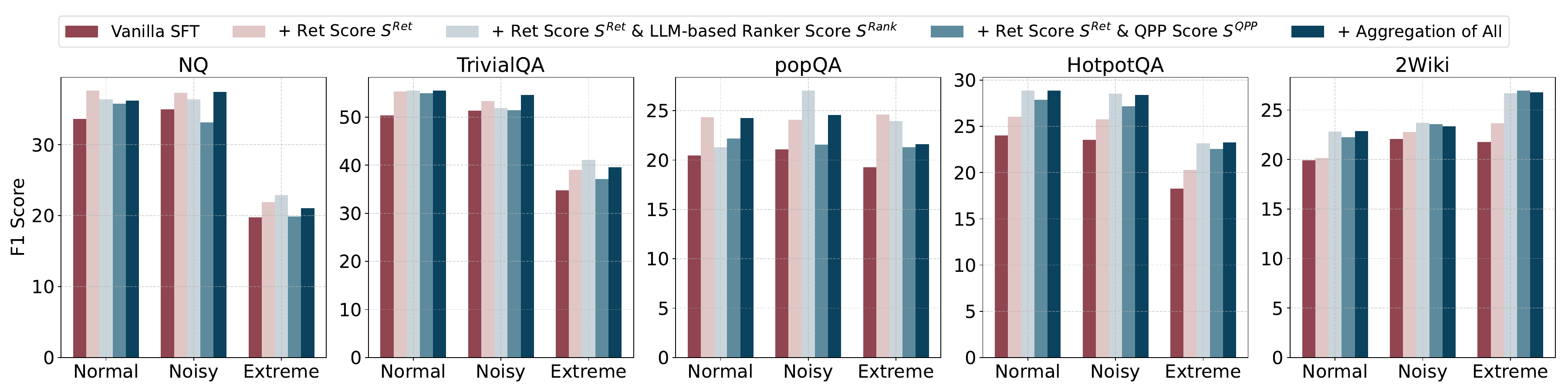}
\vspace{-2ex}
\caption{Performance of aggregating various scores as guidance features across different evaluation settings and datasets.}
\label{fig: Score_Impact}
\vspace{-1ex}
\end{figure*}
\begin{figure*}[t]
\centering
\includegraphics[width=0.97\linewidth]{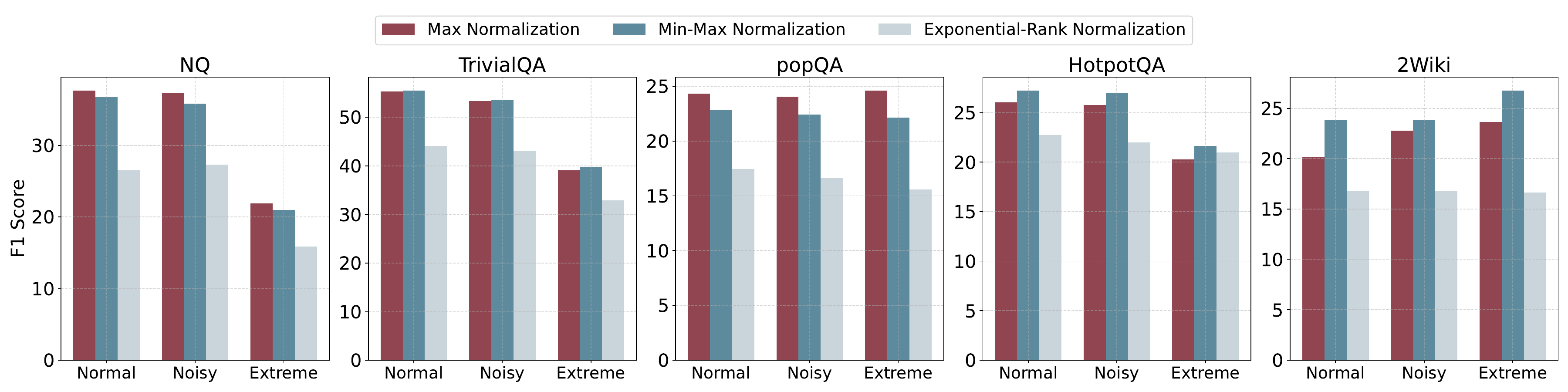}
\vspace{-2ex}
\caption{Performance of normalizing scores features with various approaches across different evaluation settings and datasets.}
\label{fig: Norm_Impact}
\vspace{-1ex}
\end{figure*}

\subsection{Feature Aggregation and Normalization}
In this section, we further investigate the impact of aggregating and normalizing various scores for answer decoding. 

\noindent \textbf{Aggregation.}
The results of score aggregation are depicted in Figure~\ref{fig: Score_Impact}. We can see that aggregating any types of relevant scores can achieve better results compared to the Vanilla SFT without explicit indicators. Leveraging the retrieval score $\mathcal{S}^{\text{Ret}}$ alone could be sufficient for the general QA datasets (NQ, TrivialQA, and popQA), where aggregating more features might not always bring additional gain. This might be because when one indicator feature is satisfied, adding the others might raise the risk of interference, as these features are measured from different aspects.
For the multi-hop QA datasets (HotpotQA and 2wiki), aggregating more feature scores helps to achieve better performance, which implies that complex questions desire more external indications to generate correct answers.
In addition, the improvement with aggregating LLM-based ranker score $\mathcal{S}^{\text{Rank}}$ compared to vanilla SFT demonstrates that the internal information processing of LLMs cannot implicitly ignore the noise, which emphasizes the importance of impacting the decoding of LLMs with explicit relevant indicators as our \ours{}.

\noindent \textbf{Normalization.}
The results of applying three normalization approaches on aggregating retrieval score $\mathcal{S}^{\text{Ret}}$ are shown in Figure~\ref{fig: Norm_Impact}. 
The Max Normalization is the simplest one, as denoted in Eq.~\ref{eq: norm}. The other two normalization approaches, Min-Max and Exponential-Rank, are implemented as
$\hat{s}_i^{\text{min-max}} = \frac{s_i - \min(\{s_j\}_{j=1}^k)}{\max(\{s_j\}_{j=1}^k) - \min(\{s_j\}_{j=1}^k)}$ and $\hat{s}_i^{\text{Exp}} = \frac{e^{-0.5(i-1)}}{\sum_{j=1}^k e^{-0.5(j-1)}}$, where the former one considers the relative gap among the original scores and the latter one further consider the impact of the rank position with exponential decay for each document candidate.
We can observe that the Max normalization performs better than the Min-Max one on general QA datasets, and vice versa on the multi-hop QA datasets. The more complex Exponential normalization with rank decay results in a large performance drop.
These observations indicate that applying different normalizations will significantly impact the performance, i.e., appropriate normalization can obtain improvement, while the inappropriate ones would result in a performance drop, even under the same pipeline in our \ours{}.
Thus, a more sophisticated approach could be further explored in future studies.

\begin{figure*}[t]
\centering
\includegraphics[width=0.97\linewidth]{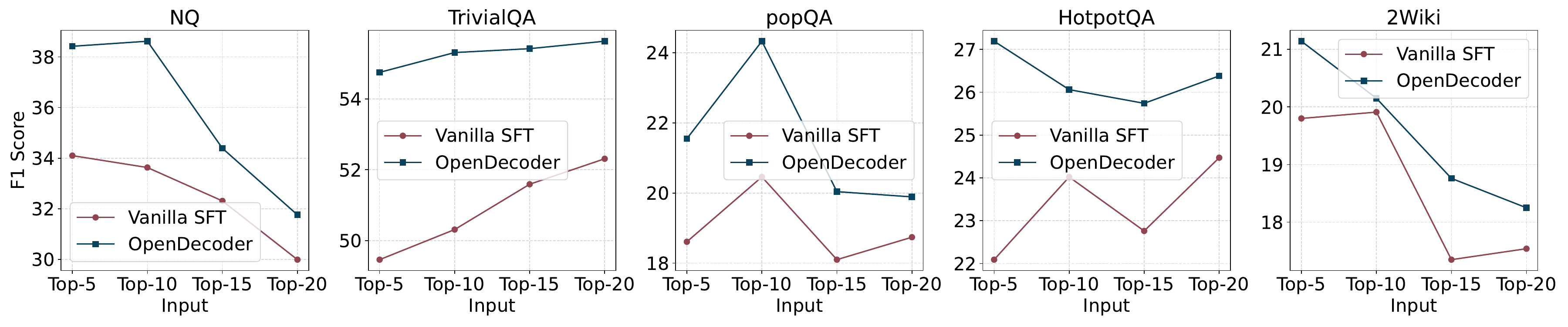}
\vspace{-2ex}
\caption{The performance of using various top-k retrieved documents in the normal evaluation setting.}
\label{fig: topk}
\vspace{-1ex}
\end{figure*}

\subsection{Document Order in Robust Training}
In this section, we examine the effect of varying document position orders on robust training.
As mentioned in Sec.~\ref{sec: Learning to Leverage Explicit Guidance Features for Decoding}, the original input context order before applying robust training is $\text{Input}=[\text{Ins.}, \text{doc}_1^q, \text{doc}_2^q, \cdots, \text{doc}_k^q, \text{q}]$. 
On top of it, we investigate three types of reorder methods, including reversing the document position from $\text{doc}_k^q$ to $\text{doc}_1^q$, shuffling them, and further injecting noise with various relevant levels as Sec.~\ref{sec: Robustness Training}.
The results are presented in Table~\ref{tab: Document Position}.
We observe that reversing the document order can obtain better performance than the original one. This might be because the new reversed order $\text{Input}^{\text{Rev.}}=[\text{Ins.}, \text{doc}_k^q, \text{doc}_{k-1}^q, \cdots, \text{doc}_1^q, \text{q}]$ enables the higher top-$k$ documents to be much closer to the question and thus might raise their attention score by alleviating the long-distance distraction. This phenomenon suggests that specifying document positions in the prompt template as plain text may not be fully interpreted by LLMs. Consequently, shuffling input documents during training can mitigate position bias, as the top-1 document is not always more informative than the top-2 for answer generation. Moreover, injecting noise further enhances model robustness by encouraging it to assess the true relevance of input documents based on external indicators, rather than relying on positional cues.

\begin{table}[t]
\caption{The performance using different document position orders in robust training across five datasets.}
\vspace{-2ex}
\label{tab: Document Position}
\centering
\scalebox{0.9}{
\begin{tabular}{lccccc}
\toprule
\multirow{1}{*}{\bf Method} & \multicolumn{1}{c}{\bf NQ} & \multicolumn{1}{c}{\bf TrivialQA} & \multicolumn{1}{c}{\bf popQA} & \multicolumn{1}{c}{\bf HotpotQA} & \multicolumn{1}{c}{\bf 2Wiki}
\\ 
\hline
Original & 35.42 & 52.57 & 20.13 & 20.26 & 22.07\\
w/. Reverse & 36.39 & 53.68 & 21.47 & 27.91 & 22.99 \\
w/. Shuffle & 37.43 & 54.57 & 24.56 & 28.39 & 23.33 \\
w/. Noise & 37.71 & 55.09 & 25.07 & 28.76 & 24.17 \\
\bottomrule
\end{tabular}}
\vspace{-2ex}
\end{table}

\subsection{Noise Tolerance of Input Top-K}
\label{sec: Noise Tolerance of Input Top-K}
As the evidence for the correct answer might relate to only a small portion of the relevant documents, the normal evaluation using the original top-$k$ retrieved results would still inevitably contain irrelevant information.
We evaluate the noise tolerance ability of Vanilla SFT and our proposed \ours{} in terms of the impact of various input top-$k$ values.
The results are shown in Figure~\ref{fig: topk}.
As the number of input documents increases, the probability of identifying relevant documents with answer information and the degree of injecting potential noise both increase.
In most of the datasets, the larger top-$k$ cannot guarantee higher performance except on TrivialQA, which indicates that the accurate search results are crucial for answer generation.
Overall, our \ours{} exhibits better performance than Vanilla SFT in different numbers of input documents, which demonstrates the effectiveness of leveraging relevance score to impact decoding across various input top-k.

\section{Conclusion}
In this paper, we propose a new paradigm to modulate the LLMs' internal information processing mechanisms with explicit indicators to improve robustness in answer decoding when the input context contains various noise.
To achieve the goal, we proposed \ours{} framework, which constructs various explicit quality indicators via extracting features from the retrieved document and applies them to modify the attention score computation among the networks of LLMs. 
Additionally, a robustness enhancement mechanism is integrated into the training procedure to enable LLMs to handle various noisy environments.
Our experiments demonstrate that incorporating explicit indicators from retrieved information in RAG tasks enhances the LLMs' ability to tolerate noise in the input context and leads to better performance compared to prior approaches.
Importantly, this paradigm is flexible to be integrated with the post-training of LLMs for any purposes and incorporated with any type of external indicators.

\begin{acks}
This work was partly supported by a discovery grant from the Natural Science and Engineering Research Council of Canada (NSERC) and the Canada Research Chair on natural language information processing and applications.
\end{acks}

\bibliographystyle{ACM-Reference-Format}
\bibliography{sample-base}

\appendix

\section*{Appendix}

\section{Datasets Details}
\label{app: Datasets}
\begin{table}[h]
\centering
\caption{Statistics of the five used datasets.}
\setlength{\tabcolsep}{4pt}{
\begin{tabular}{lrrrrr}
\toprule
 & NQ & TrivialQA & popQA & HotpotQA & 2Wiki\\ 
\midrule
\#Train Q & 79,168 & - & - & 90,447 & - \\
\#Test Q & 3,610 & 11,312 & 1,399 & 7,405 & 9,322\\
\#Collection & \multicolumn{5}{c}{21M} \\
\bottomrule
\end{tabular}}
\label{table: datasets}
\end{table}

\begin{figure*}[t]
\centering
\includegraphics[width=0.92\linewidth]{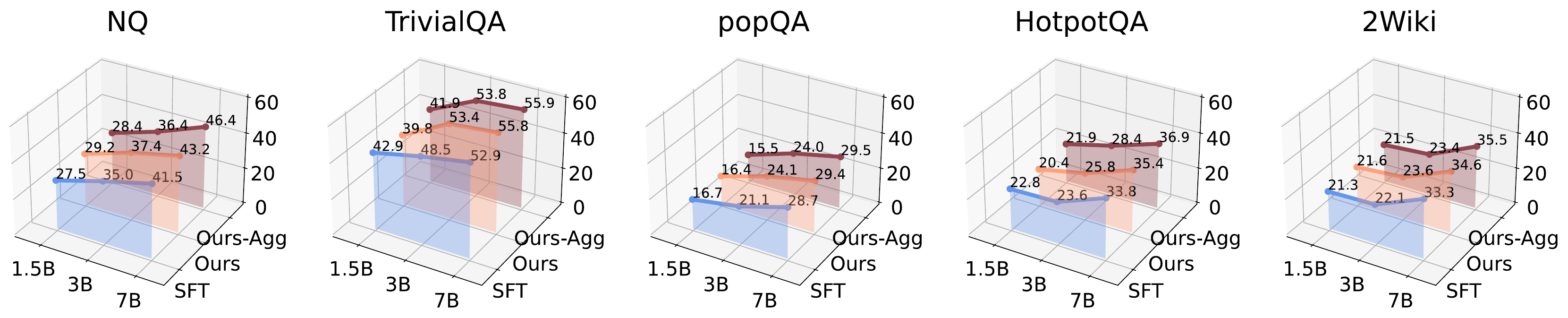}
\caption{Comparison between SFT and \ours{} of scaling model size across five datasets in the noisy evaluation setting.}
\label{fig: Scaling}
\end{figure*}
\begin{figure*}[t]
\centering
\includegraphics[width=0.9\linewidth]{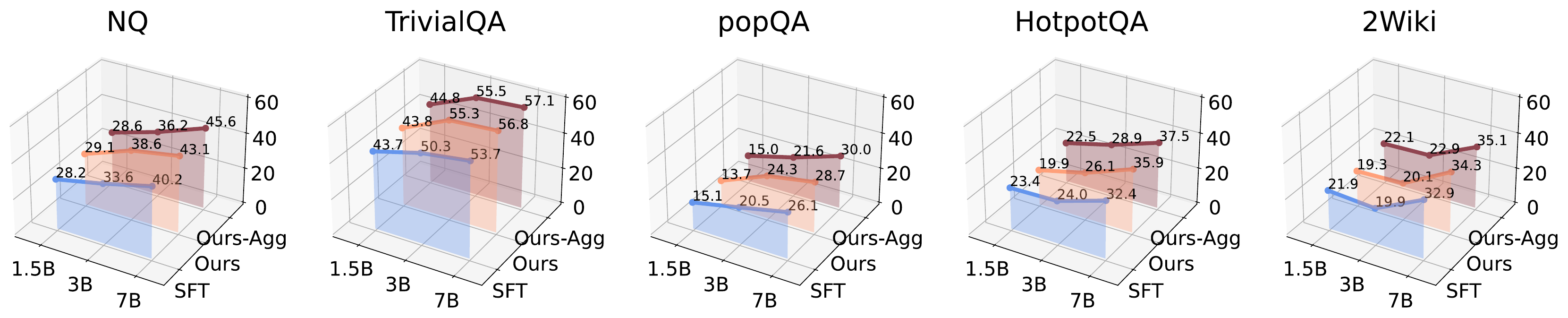}
\caption{Comparison between SFT and \ours{} of scaling model size across five datasets in the normal evaluation setting.}
\label{fig: Normal_Scaling}
\end{figure*}
\begin{figure*}[t]
\centering
\includegraphics[width=0.9\linewidth]{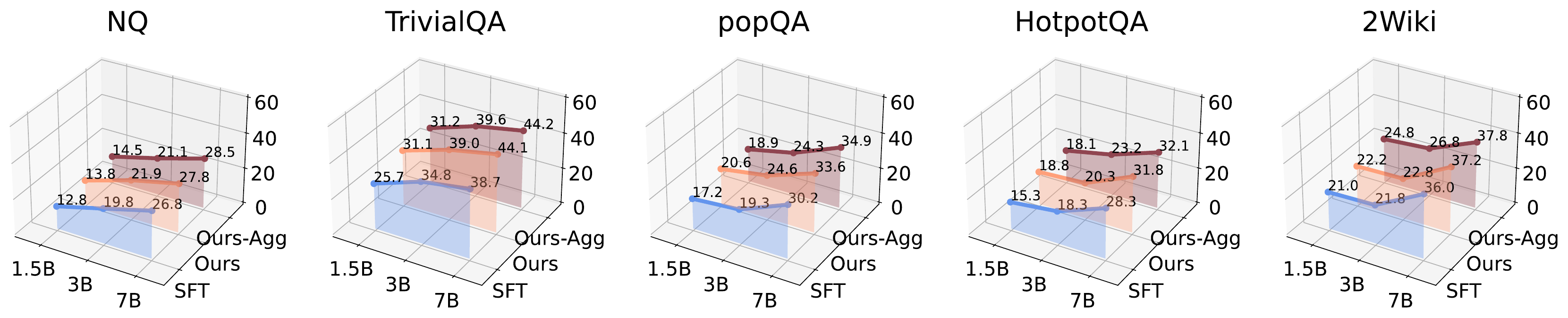}
\caption{Comparison between SFT and \ours{} of scaling model size across five datasets in the extreme noisy evaluation setting.}
\label{fig: Extreme_Scaling}
\end{figure*}

We use five benchmarks for evaluation, and the unified training set from NQ and HotpotQA to fine-tune our \ours{}. The statistics of the used datasets are presented in Table~\ref{table: datasets} and their detailed description are shown below:
\begin{itemize}
    \item \textbf{NaturalQuestion (NQ)} is a factoid dataset whose questions consist of real anonymized, aggregated queries issued to the Google search engine.
    \item \textbf{TrivialQA} is a reading comprehension dataset whose question-answer pairs authored by trivia enthusiasts and independently gathered evidence documents that provide high quality distant supervision for answering the questions.
    \item \textbf{PopQA} assesses factual question answering, challenging the model's ability to recall accurate knowledge and resolve ambiguity in entity representation.
    \item \textbf{HotpotQA} focuses on evaluating multi-hop reasoning skills, requiring models to combine information from different contexts to address a single query.
    \item \textbf{2WikiMultihopQA (2wiki)} is a dataset designed to test the model’s ability to perform multi-hop reasoning by integrating information across multiple Wikipedia passages.
\end{itemize}

\section{Baseline Details}
\label{app: Baseline}
All compared baselines are implemented by us using the same retrieved document sets across evaluation settings to guarantee fairness in comparison.
The instruction used for Vanilla RAG, Vanilla SFT, and our \ours{} is the same as ``You should answer the question by referring to the retrieved knowledge provided below and integrating the usefulness of your own parametric knowledge. Just directly answer it as a short answer without any explanation.'' 
For the prompting-based methods, RobustRAG, InstructRAG, and AstuteRAG, we inherit their original instruction provided in the corresponding code repository.
For the fine-tuning-based methods RbFT, we also use its original instruction, but set the same hyperparameter as \ours{}.

\section{Investigation of Scaling Model Size}
\label{sec: Investigation of Scaling Model Size}
We further investigate the impact of scaling up model size for vanilla SFT and our \ours{}. The results in the noisy evaluation setting are depicted in Figure~\ref{fig: Scaling}. 
Overall, both the SFT and our proposed approaches benefit from larger model sizes, suggesting that larger models are more capable of tolerating contextual noise, which aligns with prior studies~\cite{kaplan2020scaling}.
Moreover, the effectiveness of leveraging explicit indicators to influence answer generation becomes more pronounced with larger models, whereas smaller models (e.g., 1.5B) do not consistently achieve better performance across all datasets. These observations indicate that effectively integrating external signals with internal LLM reasoning processes is a non-trivial task that demands higher model capacity.
A similar trend is observed when aggregating multiple guidance score features, implying that this aggregation process also requires implicit learning during training.

The results in the normal evaluation and extreme noisy setting are also depicted in Figure~\ref{fig: Normal_Scaling} and Figure~\ref{fig: Extreme_Scaling}, respectively. 
Overall, similar trends are observed in the noisy evaluation setting, where larger models are more capable of tolerating contextual noise. Besides, the improvement in scaling model size is more pronounced in complex QA datasets than in general ones, indicating that a larger model may be equipped with a more powerful reasoning ability implicitly.
Therefore, designing more sophisticated learning objectives to better incorporate this aggregation mechanism and employing larger backbone models for training \ours{} could further enhance performance, which we leave for future work.


\section{Discussion on Time and Space Efficiency}
The computation cost of our method is the same for the offline training and online inference. 
The computation complexity of the Vanilla SFT method and our \ours{} are $\mathcal{O}(|d|^2h+|d|h^2)$ in the RAG setting, where $d$ is the average number of tokens in a document $\text{doc}$, and $h$ is the hidden dimension size of the decoder-only LLMs.
This is because the explicit guidance, i.e., the relevance scores, are produced simultaneously with the retrieved documents, and the normalization of the scores should be negligible.
In terms of the storage overhead, the normalized score $S_{\text{norm}} \in \mathbb{R}^{h \times h}$ is stored as a token-level metric, whose shape is the same as the Query, Key, and Value metric in the attention computational network inside the LLMs.
Thus, the additional storage overhead compared to Vanilla SFT is $\mathcal{O}(\text{nh})$, where $n$ is the number of Transformer layers with the impact of explicit guidance.
\end{document}